%% file: main.tex
\documentclass[11pt, a4paper, twocolumn]{article}

\usepackage[utf8]{inputenc}
\usepackage{graphicx}
\usepackage{multirow}
\usepackage{amsmath, amssymb}
\usepackage{booktabs}
\usepackage{cite}
\usepackage{hyperref}
\usepackage{geometry}
\usepackage{float}
\usepackage{color}
\usepackage{url}
\usepackage{caption}
\usepackage{subcaption}
\usepackage{algorithm}
\usepackage{algorithmic}
\usepackage{authblk}
\usepackage{newtxtext, newtxmath}
\usepackage{titlesec}
\titlespacing*{\section}{0pt}{1.5ex plus .5ex minus .2ex}{1.0ex plus .2ex}
\titlespacing*{\subsection}{0pt}{1.2ex plus .4ex minus .2ex}{0.6ex plus .1ex}
\titlespacing*{\subsubsection}{0pt}{1.0ex plus .3ex minus .2ex}{0.5ex plus .1ex}

\title{SAL: Selective Adaptive Learning for Backpropagation-Free Training with Sparsification}

\author[1,2]{Fanping Liu \thanks{Email: liufanping@rockai.net, liufanping@ruc.edu.cn}}
\author[1]{Hua Yang \thanks{Email: yanghua@rockai.net}}
\author[1]{Jiasi Zou \thanks{Email: zoujiasi@rockai.net}}

\affil[1]{ROCK AI}
\affil[2]{Renmin University of China}

\date{\today}

\begin{document}
    \maketitle

    \begin{abstract}
        Standard deep learning relies on Backpropagation (BP), which is constrained by biologically implausible weight symmetry and suffers from significant gradient interference within dense representations. To mitigate these bottlenecks, we propose \textbf{Selective Adaptive Learning (SAL)}, a training method that combines selective parameter activation with adaptive area partitioning. Specifically, SAL decomposes the parameter space into mutually exclusive, sample-dependent regions. This decoupling mitigates gradient interference across divergent semantic patterns and addresses explicit weight symmetry requirements through our refined feedback alignment. Empirically, SAL demonstrates competitive convergence rates, leading to improved classification performance across 10 standard benchmarks. Additionally, SAL achieves numerical consistency and competitive accuracy even in deep regimes (up to 128 layers) and large-scale models (up to 1B parameters). Our approach is loosely inspired by biological learning mechanisms, offering a plausible alternative that contributes to the study of scalable neural network training.\footnote{Our code is available at \href{https://anonymous.4open.science/r/icml26_SAL-37CE/}{this repository}.}

    \end{abstract}

    \section{Introduction}

    The core training mechanism of deep neural networks, the Backpropagation (BP) algorithm, has long been criticized for its biological implausibility, particularly the "weight transport problem" \cite{Lillicrap2016Random}. To explore alternative methods, representative attempts such as Feedback Alignment (FA) and Direct Feedback Alignment (DFA) bypass the weight symmetry requirement by employing fixed random feedback matrices \cite{Lillicrap2016Random, Nokland2016Direct}. More recently, these challenges have been further formalized regarding resource efficiency and global synchronization \cite{Lillicrap2020Backpropagation}. In response to these limitations, Hinton proposed the Forward-Forward (FF) algorithm, which achieves layer-wise decoupling through local goodness maximization \cite{Hinton2022Forward}. However, these backpropagation-free (BP-free) methods still exhibit a performance gap compared to BP when applied to deep, large-scale models \cite{Filipovich2022Scaling}.

    To reduce resource consumption and improve training efficiency, Dynamic Sparse Training (DST) methods, such as SET and RigL, adjust connectivity in real-time during training, thereby avoiding the high initial costs associated with traditional pruning \cite{Han2016DeepCompression, Mocanu2018Scalable, Evci2020Rigging}. Concurrently, the selective activation and sparse update characteristics of biological neurons \cite{Roy2019Towards} have inspired various selective learning mechanisms. Nevertheless, research that systematically integrates selectivity, adaptivity, and sparsity within a BP-free paradigm remains scarce.

    In this paper, we address these limitations by combining selective parameter activation with adaptive area partitioning. In conventional setups, distinct data patterns share the entire parameter space, which can lead to conflicting gradient updates in which the optimization for one semantic cluster may inadvertently degrade the representation of another. This issue is exacerbated in BP-free approaches, such as FA, where the absence of precise gradient information renders deep layers increasingly susceptible to noise accumulation \cite{Li2024Deep}. To circumvent this limitation, we integrate contemporary neuroscientific insights regarding the functional specialization of the cerebral cortex \cite{Du2024Organization}. By adopting this principle of routing inputs to specialized sub-regions, we aim to ensure that synaptic updates remain locally relevant and globally coherent, which contributes to stabilizing the learning dynamics without relying on end-to-end signal propagation.

    We introduce \textbf{Selective Adaptive Learning (SAL)}, a biologically inspired training method. Unlike Dynamic Sparse Training (DST), which sparsifies weights post-hoc or during training based on magnitude, SAL employs a \textit{Learned-Frozen Decoupled Routing} strategy. This mechanism dynamically partitions the input space into mutually exclusive areas, directing samples to specialized parameter subspaces. By strictly isolating the activation paths, SAL potentially reduces gradient interference and allows for the use of fixed, asymmetric feedback connections mitigating the stability issues typically associated with deep FA networks.

    Our contributions are summarized as follows:

    \begin{itemize}
        \item We propose a learnable, adaptive per-layer routing strategy that achieves explicit parameter space decoupling. This enables scaling model capacity with limited additional computational cost, as only a single area is activated and updated per sample, thereby promoting global sparsity throughout the network.
        \item By integrating an asymmetric error propagation scheme with local learning signals, the model circumvents the weight symmetry requirement of backpropagation. This allows each layer to optimize against localized alignment objectives, facilitating stable convergence in deep hierarchies. The resulting update rules are layer-wise decoupled and do not impose inter-layer synchronization, thereby structurally permitting parallel parameter updates in principle.
        \item We provide empirical evidence demonstrating that SAL outperforms or remains competitive with standard BP baselines across several benchmarks. Notably, SAL scales effectively to both deep regimes (up to 128 layers) and large-scale models (up to 1 billion parameters).
    \end{itemize}

    \section{Related Work}

    The SAL integrates bio-inspired local learning with BP-free and learnable routing training. By synthesizing sparse, area-conditional computation with a form of asymmetric error propagation, this work is situated at the intersection of the aforementioned research areas.

    \subsection{BP-Free and Biologically Plausible Learning}

    To address the biological implausibility of BP---specifically weight transport and update locking---several alternatives have emerged. FA and its derivatives, such as DFA, demonstrate that fixed random matrices can effectively propagate error signals through self-organizing alignment \cite{Lillicrap2016Random, Nokland2016Direct, Refinetti2021Align}. Recent architectures like Direct Random Target Projection (DRTP) \cite{Frenkel2021Learning}, Product Feedback Alignment (PFA) \cite{Li2024Deep}, and Feedforward Projection (FP) \cite{OShea2025Closed} further relax these constraints by minimizing feedback connectivity requirements.

    Alternative approaches include Target Propagation (TP), which optimizes hidden layers via target states rather than gradients \cite{Bengio2014How, Lee2015Difference}. More recently, the FF algorithm introduced layer-wise ``goodness'' optimization \cite{Hinton2022Forward}, a method that has been extended to Spiking Neural Networks (SNNs) to enhance alignment with biological principles \cite{Ghader2025Backpropagation}. In this context, the SAL method aims to balance biological plausibility with model stability by integrating global error signals with local alignment strategies.

    \subsection{Sparse Training and Dynamic Mechanisms}

    Weight sparsity is an effective technique for alleviating computational bottlenecks and improving model generalization in various settings. Prior work, such as Sparse Evolutionary Training (SET) \cite{Mocanu2018Scalable, Mostafa2019Parameter, Evci2019Difficulty} and Dynamic Sparse Training (DST) variants—including RigL \cite{Evci2020Rigging} and SRigL \cite{Lasby2024Dynamic}—demonstrate that networks can maintain competitive performance by dynamically adapting topologies during training. These empirical improvements align with theoretical frameworks like Bregman Sparse Learning \cite{Bungert2022Bregman}.

    Sparsity also plays a key role in shaping learning dynamics. In FA regimes, sparse feedback enhances scalability for deep architectures \cite{Crafton2019Direct, Launay2020Direct}, while in continual learning, it mitigates catastrophic forgetting by isolating task-specific parameters \cite{Lassig2022Bioinspired}. Additionally, Mixture of Experts (MoE) models \cite{Shazeer2017Outrageously, Lepikhin2021GShard} utilize conditional computation to scale model capacity within fixed computational budgets.

    Building on these principles, the SAL method employs layer-wise dynamic adaptive selection, resulting in inherent model sparsity.

    \subsection{Bio-inspired and Local Learning}

    Selective learning mechanisms observed in cortical circuits have motivated a wide range of computational models focusing on meta-learning and local plasticity rules \cite{Lindsey2020Learning,Payeur2021Burstdependent, Lassig2022Bioinspired, Shervani2023Meta}. Moreover, predictive coding and hierarchical local target learning \cite{Whittington2017Approximation} have further contributed to the development of BP-free optimization methods by emphasizing local error minimization mechanisms.

    While existing methods, such as Difference Target Propagation and predictive coding \cite{Bengio2015Towards, Whittington2017Approximation}, provide theoretical or empirical foundations, they often treat sparsity and BP-free learning as disjoint objectives. The proposed SAL method bridges this gap by integrating multi-layer local learning with a selective adaptive parameter mechanism.

    \section{Methodology}

    \subsection{Overview of the Proposed Method}

\begin{figure*}[t]
        \centering
        \includegraphics[width=0.8\textwidth]{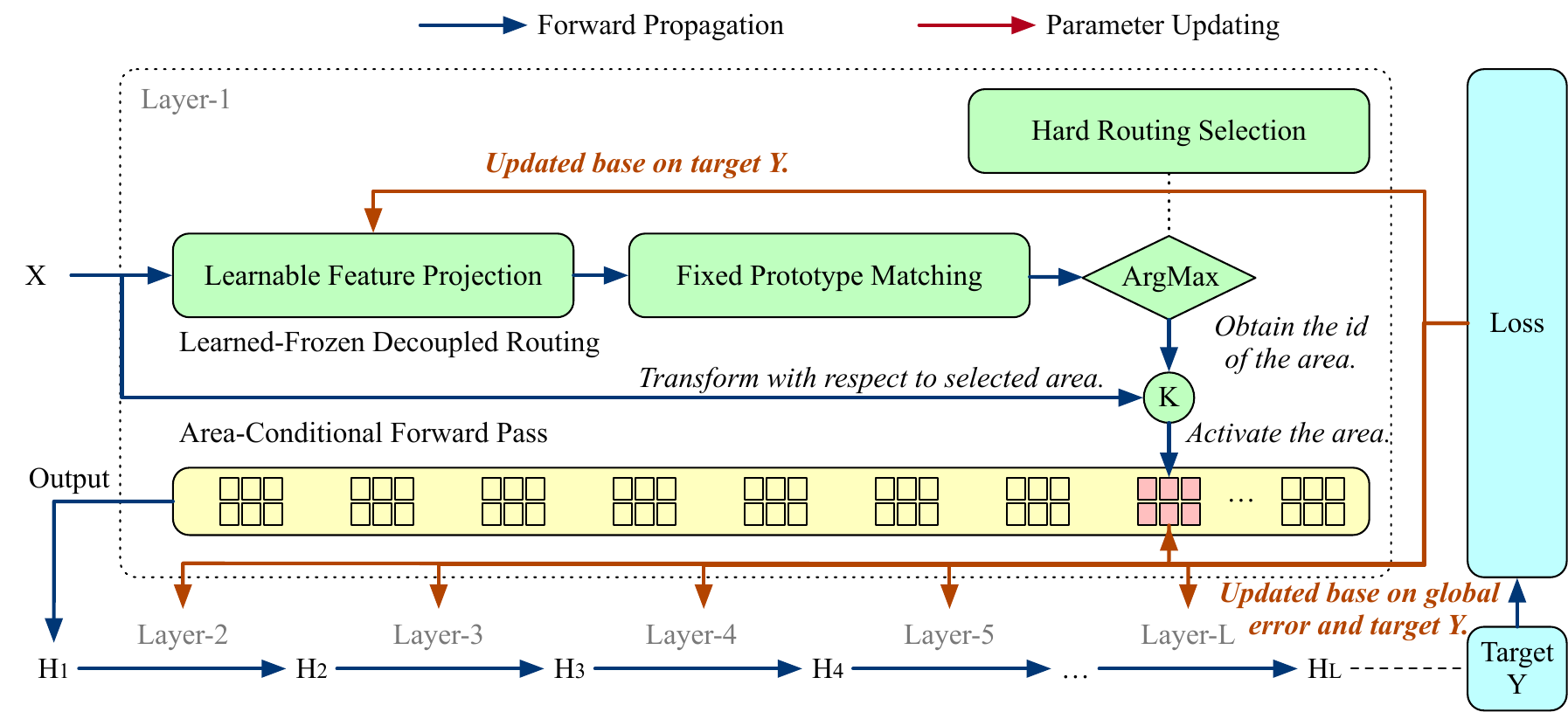}
        \caption{A multi-layer network architecture based on SAL.}
        \label{fig:A multi-layer network architecture based on SAL}
    \end{figure*}

    The SAL method combines selective parameter activation with adaptive area partitioning. It comprises two separate phases: forward propagation and parameter updating; notably, the latter deviates from the standard backpropagation process. The multi-layer network structure based on SAL is illustrated in Figure~\ref{fig:A multi-layer network architecture based on SAL}. Its primary objective is to maintain a per-sample forward computational complexity equivalent to that of standard feedforward layers while reducing gradient interference and improving stability through explicit parameter space decoupling.

    In conventional fully-connected (FC) layers, the linear transformation with global connectivity $\mathbf{h} = \phi(\mathbf{x}\mathbf{W} + \mathbf{b})$ assumes a relatively homogeneous data distribution. However, practical high-dimensional data often reside on diverse low-dimensional sub-manifolds. Constraining a single set of parameters to fit these distinct semantic clusters can potentially lead to gradient interference: (i) oscillatory convergence, (ii) biased overfitting toward dominant sub-distributions, and (iii) poor representation of long-tail patterns.

    SAL incorporates area adaptation to partition the parameter space into mutually exclusive areas. By restricting each input to a single activated subspace, SAL enforces sample-wise parameter isolation. This design is inspired by two biological principles of biological neural systems:

    \begin{itemize}
        \item \textbf{Sparsity:} Only a sparse subset of neurons is activated during specific cognitive tasks, a phenomenon that is consistently reported across the cortex\cite{Barth2012Experimental,Jaaskelainen2022Do}.

        \item \textbf{Alignment:} Synaptic weight updates do not necessitate strict symmetry between forward and backward passes; instead, error signals can be propagated via asymmetric feedback pathways\cite{Li2024Deep,Khan2025Predictive}.
    \end{itemize}

    Building on these biological insights, SAL aims to translate the principles of sparsity and alignment into a structured computational method. To describe the proposed method, we first introduce single-layer forward propagation, then extend it to the multi-layer computations, and finally present the mechanisms for error propagation and parameter updates.

    \subsection{Single-Layer Forward Propagation}

    Consider a single-layer neural module. Let the input vector be $\mathbf{x} \in \mathbb{R}^{d_{in}}$ and the output vector be $\mathbf{h} \in \mathbb{R}^{d_{out}}$.

    \subsubsection{Learned-Frozen Decoupled Routing}

    The routing mechanism is designed to maintain learnability while preventing mode collapse (e.g., the convergence of all samples to a single region in the representation space). SAL employs a ``Learned-Frozen Decoupling'' strategy, which decomposes the routing process into three distinct stages.

    \paragraph{Learnable Feature Projection}
    First, the input is projected into a routing feature space via a learnable linear mapping:
    \begin{equation}
        \mathbf{z} = \mathbf{x}\mathbf{W}_s,
        \quad
        \mathbf{W}_s \in \mathbb{R}^{d_{in} \times d_f},
    \end{equation}
    where $d_f$ denotes the routing feature dimension. To minimize the additional computational overhead, $d_f = d_{out}$ is typically set for a single-layer configuration.

    \paragraph{Fixed Prototype Matching}
    Second, we define a set of projection matrices that are randomly initialized and kept frozen during training:
    \[
        \mathbf{W}_{fix} \in \mathbb{R}^{d_f \times N},
    \]
    where $N$ represents the number of areas. $\mathbf{W}_{fix}$ provides $N$ fixed directional anchors in the routing feature space. Matching scores are computed as:
    \begin{equation}
        \mathbf{p} = \mathbf{z}\mathbf{W}_{fix}.
    \end{equation}
    Since $\mathbf{W}_{fix}$ does not participate in updates, routing diversity is driven entirely by the learning behavior of $\mathbf{W}_s$, effectively preventing routing degradation into a single area.

    \paragraph{Hard Routing Selection}
    Third, the final area index is determined via a hard selection process:
    \begin{equation}
        k = \operatorname*{arg\,max}_{j \in \{1,\dots,N\}} p_j.
    \end{equation}
    This hard routing mechanism ensures that each sample activates exactly one area during the forward pass, thereby achieving sparse activation.

    \subsubsection{Area-Conditional Forward Pass}

    The SAL maintains a collection of area-conditional parameter tensors:
    \[
        \mathcal{W} \in \mathbb{R}^{N \times d_{in} \times d_{out}}, \quad
        \mathcal{B} \in \mathbb{R}^{N \times d_{out}}.
    \]
    Given the routing index $k$, only the $k$-th set of parameters is selected for computation:
    \[
        \mathbf{u} = \mathbf{x}\mathbf{W}^{(k)} + \mathbf{b}^{(k)},
    \]
    \begin{equation}
        \mathbf{h} = \phi(\mathbf{u}),
    \end{equation}
    Given that each sample performs only one linear mapping, the forward computational complexity remains $O(d_{in} d_{out})$, which is consistent with standard FC layers.

    \subsection{Multi-Layer Forward Computation}

    A deep neural network can be constructed by stacking $L$ SAL layers. We define $\mathbf{h}^{(0)} = \mathbf{x}_{input}$ and $\mathbf{x}^{(l)} = \mathbf{h}^{(l-1)}$. When input and output dimensions match, SAL supports residual connections to mitigate the gradient vanishing and facilitate identity mapping:
    \begin{equation}
        \mathbf{h}^{(l)} =
        \phi\!\left(
                  \mathbf{x}^{(l)}\mathbf{W}^{(l,k^{(l)})}
                  + \mathbf{b}^{(l,k^{(l)})}
        \right)
        + \mathbf{x}^{(l)}.
    \end{equation}
    Routing selection is performed independently at each layer, allowing the network to learn hierarchical area-partitioning strategies.

    \subsection{Error Propagation and Parameter Update}

    \subsubsection{Asymmetric Error Propagation Path}

    SAL introduces a refined alignment mechanism. For the $l$-th layer, a fixed feedback matrix is defined:
    \[
        \mathbf{B}^{(l)} \in \mathbb{R}^{d_{out} \times d_f}.
    \]
    If $d_{out} = d_f$, $\mathbf{B}^{(l)}$ is typically set as an identity matrix (common in the output layer); otherwise, it is initialized randomly and remains frozen during training. Let $\mathbf{E}_{top}$ be the global error signal from the output layer. The effective error signal for the current layer is:
    \begin{equation}
        \boldsymbol{\delta}^{(l)}
        =
        \phi'(\mathbf{u}^{(l)})
        \odot
        \left(
            \mathbf{E}_{top}
            +
            \mathbf{E}_{local}^{(l)}
        \right),
    \end{equation}
    where $\mathbf{E}_{local}^{(l)}$ is generated via the fixed feedback pathway to provide local supervision.

    \subsubsection{Area-Conditional Parameter Update}

    Only the areas activated within the current mini-batch undergo parameter updates. Let $S_k$ denote the set of samples activating area $k$:
    \begin{equation}
        \nabla \mathbf{W}^{(k)} =
        \sum_{i \in S_k}
        \mathbf{x}_i^\top \boldsymbol{\delta}_i,
    \end{equation}
    \begin{equation}
        \nabla \mathbf{b}^{(k)} =
        \sum_{i \in S_k}
        \boldsymbol{\delta}_i.
    \end{equation}
    This method ensures that gradients for different areas are decoupled within the parameter space.

    \subsubsection{Optimizer Strategy for the Selector}

    Since the $\arg\max$ operation is non-differentiable, SAL decouples the selector's training from the backbone updates. For each layer, we define an auxiliary loss function aligned with the primary task (e.g., Cross-Entropy for classification):

    \begin{equation}
        \mathcal{L}_{sel}^{(l)} =
        \text{CrossEntropy}
        \left(
            \text{Softmax}(\mathbf{z}^{(l)}),
            \mathbf{y}_{target}
        \right).
    \end{equation}
    Selector parameters $\mathbf{W}_s^{(l)}$ are updated via this auxiliary loss independently of the main network's gradients, directing the routing strategies toward task-relevant semantics.

    \subsection{Detailed Algorithms}

    Algorithm~\ref{alg:complete_sal_training} outlines the operational logic of the SAL-based multi-layer neural network and details the methods described above.

    \begin{algorithm}[!t]
        \caption{The SAL Training Procedure}
        \label{alg:complete_sal_training}
        \begin{algorithmic}[1]
            \REQUIRE
            Dataset $\mathcal{D} = \{(\mathbf{X}_m, \mathbf{Y}_m)\}$, total epochs $T$, batch size $B$, number of layers $L$

            \REQUIRE
            For each layer $l \in [1, L]$:
            Learnable $\{\mathbf{W}_s^{(l)}, \mathbf{W}^{(l,k)}, \mathbf{b}^{(l,k)}\}$,
            Fixed parameters $\{\mathbf{W}_{fix}^{(l)}, \mathbf{B}^{(l)}\}$

            \STATE Initialize all learnable parameters $\Theta$ randomly

            \FOR{epoch $= 1$ to $T$}
            \FOR{each mini-batch $(\mathbf{X}, \mathbf{Y}) \in \mathcal{D}$}

            \STATE \textbf{--- 1. Multi-layer Forward Phase ---}
            \STATE $\mathbf{H}^{(0)} \leftarrow \mathbf{X}$
            \FOR{$l = 1$ to $L$}
            \STATE \textit{// Step A: Routing}
            \STATE $\mathbf{Z}^{(l)} \leftarrow \mathbf{H}^{(l-1)} \mathbf{W}_s^{(l)}$
            \STATE $\mathbf{P}^{(l)} \leftarrow \mathbf{Z}^{(l)} \mathbf{W}_{fix}^{(l)}$
            \STATE $K_i^{(l)} \leftarrow \arg\max_j \mathbf{P}_{i,j}^{(l)}$ for $i \in \{1,\dots,B\}$

            \STATE \textit{// Step B: Area-specific Activation}
            \FOR{each active area $k$ in layer $l$}
            \STATE $I_k \leftarrow \{ i \mid K_i^{(l)} = k \}$
            \STATE $\mathbf{U}^{(l)}[I_k] \leftarrow \mathbf{H}^{(l-1)}[I_k]\mathbf{W}^{(l,k)} + \mathbf{b}^{(l,k)}$
            \STATE $\mathbf{H}^{(l)}[I_k] \leftarrow \phi(\mathbf{U}^{(l)}[I_k])$
            \ENDFOR
            \ENDFOR

            \STATE \textbf{--- 2. Error Computation ---}
            \STATE Compute Global Loss: $\mathcal{L}_{total} = \mathcal{L}(\mathbf{H}^{(L)}, \mathbf{Y})$
            \STATE Global Output Error: $\mathbf{E}_{top} \leftarrow \nabla_{\mathbf{H}^{(L)}} \mathcal{L}_{total}$

            \STATE \textbf{--- 3. Update Phase ---}
            \FOR{$l = 1$ to $L$}
            \STATE \textit{// A. Update Selector Layer}
            \STATE Compute $\mathcal{L}_{sel}^{(l)}(\mathbf{Z}^{(l)}, \mathbf{Y})$ \COMMENT{Task-aligned auxiliary loss}
            \STATE $\mathbf{W}_s^{(l)} \leftarrow \mathbf{W}_s^{(l)} - \eta_{sel} (\mathbf{H}^{(l-1)\top} \nabla_{\mathbf{Z}^{(l)}} \mathcal{L}_{sel}^{(l)})$

            \STATE \textit{// B. Update Area Parameters}
            \STATE $\mathbf{E}_{local}^{(l)} \leftarrow \nabla_{\mathbf{H}^{(l)}} \mathcal{L}_{local}(\mathbf{H}^{(l)}\mathbf{B}^{(l)}, \mathbf{Y})$
            \STATE $\mathbf{E}^{(l)} \leftarrow \mathbf{E}_{top} + \mathbf{E}_{local}^{(l)}$
            \STATE $\boldsymbol{\delta}^{(l)} \leftarrow \phi'(\mathbf{U}^{(l)}) \odot (\mathbf{E}^{(l)}\mathbf{B}^{(l)\top})$

            \FOR{each active area $k$ in layer $l$}
            \STATE $\nabla \mathbf{W}^{(l,k)} \leftarrow \mathbf{H}^{(l-1)}[I_k]^\top \boldsymbol{\delta}^{(l)}[I_k]$
            \STATE $\mathbf{W}^{(l,k)} \leftarrow \mathbf{W}^{(l,k)} - \eta_{net} \nabla \mathbf{W}^{(l,k)}$
            \STATE $\mathbf{b}^{(l,k)} \leftarrow \mathbf{b}^{(l,k)} - \eta_{net} \sum \boldsymbol{\delta}^{(l)}[I_k]$
            \ENDFOR

            \ENDFOR

            \ENDFOR
            \ENDFOR
        \end{algorithmic}
    \end{algorithm}

    \subsection{Computational Complexity}

    We analyze the efficiency of SAL relative to a standard FC layer with dimension $D$, assuming that both the input and output have dimensionality $D$. Let $N$ be the number of areas and $d_f$ be the routing feature dimension.

    \paragraph{Space Complexity.}
    A standard FC layer requires $O(D^2)$ parameters. SAL maintains $N$ mutually exclusive areas, leading to $O(ND^2)$ weight parameters. The routing mechanism introduces $O(D d_f)$ additional parameters; however, since $d_f \leq D$, this overhead remains negligible relative to the area weights.

    \paragraph{Time Complexity.}
    The per-sample computational cost is summarized as follows:
    \begin{itemize}
        \item \textbf{Forward Pass:} SAL computes routing scores in $O(Dd_f)$ and performs a single area transformation in $O(D^2)$. The total complexity $O(D^2 + Dd_f)$ is asymptotically equivalent to a standard FC layer ($O(D^2)$) given $d_f \approx D$.
        \item \textbf{Update Pass:} While SAL contains more parameters, the error signal $\boldsymbol{\delta}$ only triggers updates for a single active area $\mathbf{W}^{(k)}$. Consequently, the number of floating-point operations (FLOPs) for gradient computation and weight updates per sample is comparable to that of a standard BP-trained layer.
    \end{itemize}

    \section{Experiments}

    To assess the empirical performance and generalization capability, we evaluated the proposed SAL method across ten public datasets. These benchmarks represent a wide spectrum of visual recognition tasks, including natural object classification, medical pathology imaging, and handwritten character recognition. Detailed statistics and preprocessing schemes for each dataset are summarized in Table~\ref{tab:datasets}.

    \begin{table*}[t]
        \centering
        \caption{For the Digits dataset, we apply standard Z-score normalization (transforming the data to a distribution with a mean of $0$ and a standard deviation of $1$). For all other datasets, pixel values are normalized using a mean of $0.5$ and a standard deviation of $0.5$, resulting in values in the range $[-1, 1]$.}
        \label{tab:datasets}
        \scshape
        \begin{tabular}{lccccc}
            \toprule
            Dataset                                 & Classes & Total Samples & Resolution     & Channel Processing     \\
            \midrule
            CIFAR-10 \cite{Krizhevsky2009Learning}  & 10      & 60,000        & $32 \times 32$ & RGB $\rightarrow$ Gray \\
            PCam \cite{Veeling2018Rotation}         & 2       & 327,680       & $96 \times 96$ & RGB $\rightarrow$ Gray \\
            STL-10 \cite{Coates2011Analysis}        & 10      & 13,000        & $96 \times 96$ & RGB $\rightarrow$ Gray \\
            SVHN \cite{Netzer2011Reading}           & 10      & 99,289        & $32 \times 32$ & RGB $\rightarrow$ Gray \\
            MNIST \cite{LeCun1998Gradient}          & 10      & 70,000        & $28 \times 28$ & Gray                   \\
            Fashion-MNIST \cite{Xiao2017Fashion}    & 10      & 70,000        & $28 \times 28$ & Gray                   \\
            Digits \cite{Alpaydin1998Optical}       & 10      & 1,797         & $8 \times 8$   & Gray                   \\
            USPS \cite{Hull1994Database}            & 10      & 9,298         & $16 \times 16$ & Gray                   \\
            Semeion \cite{Buscema1998Semeion}       & 10      & 1,593         & $16 \times 16$ & Gray                   \\
            FER2013 \cite{Goodfellow2013challenges} & 7       & 35,887        & $48 \times 48$ & Gray                   \\
            \bottomrule
        \end{tabular}
    \end{table*}

    The primary objective of our experiments is not to achieve state-of-the-art (SOTA) performance on each dataset, but rather to provide a controlled comparison between the proposed SAL method and a standard BP baseline using identical network architectures.

    \subsection{Basic Convergence Validation}
    \label{lab:basic_convergence_validation}

    \subsubsection{Experimental Setup}

    We first evaluated the convergence of the model in a shallow architecture using a two-layer fully connected network. The hidden layer size was fixed at 256, while the input and output dimensions were determined by the specific dataset (e.g., $784 \rightarrow 10$ for MNIST). We employed ReLU activation with Kaiming initialization\cite{He2015Delving}. Both the Baseline and SAL models were trained for 25 epochs using SGD with a batch size of 16 and a learning rate of 0.0001. All results were averaged over five independent runs with different random seeds.

    For the SAL method, we varied the number of areas $n_{\text{areas}} \in \{1, 2, 4, 8, 16\}$ in the first layer to evaluate its impact relative to the baseline.

    \subsubsection{Results}

    As shown in Table~\ref{tab:main_final_test_accuracy_sal_vs_baseline}, SAL outperforms the baseline on nine out of the ten datasets. For example, on the Digits dataset, accuracy increases from 38.53\% (Baseline) to 71.63\% (SAL-16), representing a 33.1 percentage point improvement. On Semeion, SAL-16 achieves 72.03\%, marking a notable improvement of 36.87 percentage points over the baseline's 35.16\%. These results suggest that the SAL method effectively captures intrinsic data features, consistently enhancing the model's representation and classification capabilities.

    In addition, a positive correlation or a stable high-performance trend was observed across datasets such as the CIFAR-10, Digits, FER2013, Fashion-MNIST, MNIST, STL-10, Semeion, and USPS as $n_{\text{areas}}$ increases. This suggests that tuning the $n_{\text{areas}}$ parameter allows the model to effectively leverage its structural capacity.

    \begin{table*}[t]
        \caption{Comparison of test accuracies between SAL and baselines across 10 benchmarks. The table reports the mean accuracy (in \%) at the final epoch, averaged over five runs with different random seeds.}
        \label{tab:main_final_test_accuracy_sal_vs_baseline}
        \begin{center}
            \begin{small}
                \begin{sc}
                    \input{figures/main_final_test_accuracy_sal_vs_baseline.tex}
                \end{sc}
            \end{small}
        \end{center}
        \vskip -0.1in
    \end{table*}

    \subsection{Impact of Network Depth}
    \label{lab:impact_of_network_depth}
    To further investigate the performance across different depths, we selected four representative datasets with varying feature dimensions: CIFAR-10, FER2013, Semeion, and STL-10.

    \subsubsection{Experimental Setup}
    The architectures were kept consistent between the baseline and SAL. We evaluated depths of $\{4, 16, 64, 128\}$ with a hidden layer dimension of 256. The tanh activation function was applied to all layers except for the first and the final layers, which used the linear activation function. For SAL, $n_{\text{areas}}$ was set to 4 for both the input and all hidden layers. To facilitate stable gradient flow, we employed residual connections across the hidden layers. Other training settings and hyperparameters were identical to those in Section~\ref{lab:basic_convergence_validation}.

    \subsubsection{Results}
\begin{figure*}[t]
        \centering
        \includegraphics[width=0.8\textwidth]{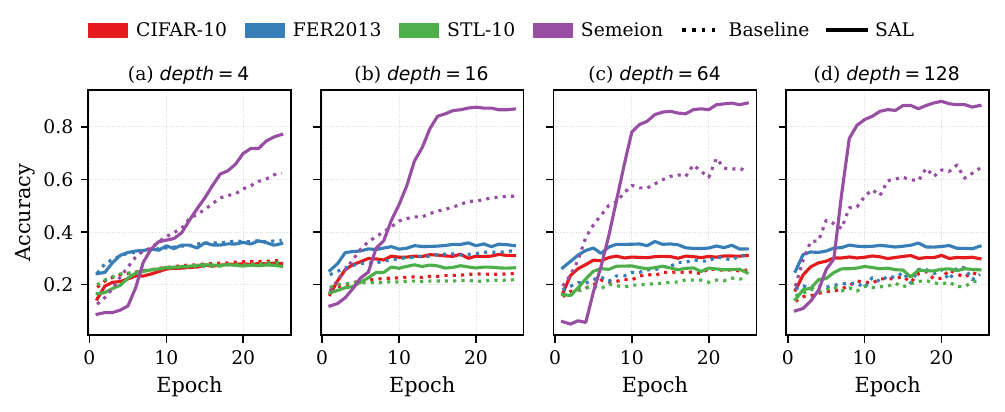}
        \caption{Performance comparison between SAL and the baseline across various network depths on 4 datasets. Across all plots, a consistent encoding is used: colors distinguish datasets, and line styles differentiate methods (e.g., solid for SAL, dotted for baseline).}
        \label{fig:main_depth_scalability_on_4_datasets}
    \end{figure*}

    As shown in Figure~\ref{fig:main_depth_scalability_on_4_datasets}, the performance gap between SAL and the baseline increases with network depth. While the baseline undergoes degradation and optimization difficulties at greater depths (e.g., $depth=128$), SAL remains comparatively stable.

    Using the Semeion dataset as a benchmark, the BP model demonstrates steady performance improvements. Concurrently, the SAL method exhibits characteristics typical of traditional neural networks, with its performance scaling in proportion to model depth.

    \subsection{Impact of Network Width}
    \label{lab:impact_of_network_width}
    This experiment further examined the impact of model width on CIFAR-10, FER2013, Semeion, and STL-10 by scaling the hidden layer dimensions.

    \subsubsection{Experimental Setup}

    Fixing the depth at 64 layers, we set the hidden layer sizes to $\{1024, 2048, 4096\}$. For SAL, $n_{\text{areas}}$ was set to 4 for the input layer and all hidden layers. All other training configurations remained consistent with Section~\ref{lab:impact_of_network_depth}. The parameter scale expanded with the hidden layer dimension; specifically, at a width of 4096, the SAL method involved approximately 1B activation parameters and over 4B total parameters.

    \subsubsection{Results}

\begin{figure*}[!t]
        \centering
        \includegraphics[width=0.8\textwidth]{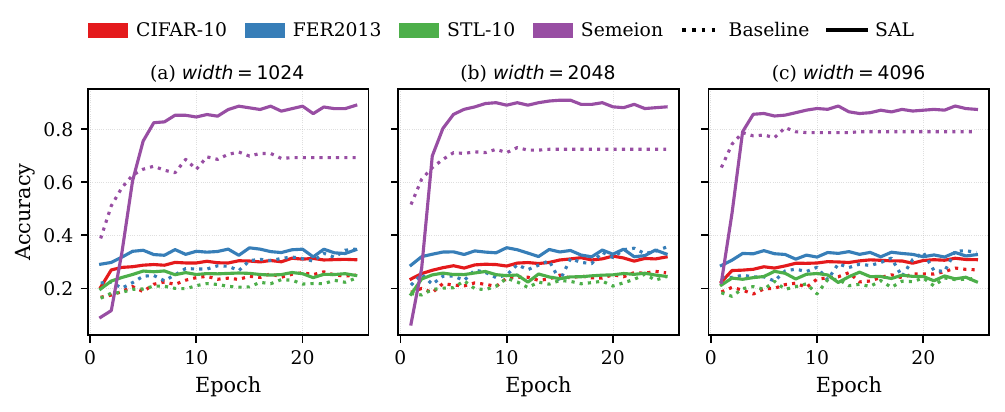}
        \caption{Performance comparison between SAL and the baseline across various network widths on 4 datasets.}
        \label{fig:main_width_scalability_on_4_datasets}
    \end{figure*}

    Figure~\ref{fig:main_width_scalability_on_4_datasets} shows the accuracy trends for the baseline and SAL on 4 datasets.

    The SAL method consistently improves the convergence upper bound and learning efficiency across various network widths. On specific datasets such as Semeion, the proposed method yields notable performance gains over the baseline and highly adaptable to the underlying data structures.

    \section{Discussion}

    This section discusses the key architectural properties of SAL, focusing on its comparative advantages over MoE, its convergence stability, and current limitations.

    \subsection{Comparative Analysis: SAL vs. MoE}
    \label{sec:sal_vs_moe}

    While SAL and MoE both utilize sparsity to optimize computation, they operate on fundamentally different principles. MoE is primarily a scaling strategy designed to increase model capacity through conditional computation. In contrast, SAL is a learning paradigm centered on layer-wise decoupling and asymmetric alignment.

    Experimental results in Figure~\ref{fig:main_sal_with_moe_on_4_datasets} demonstrate that SAL achieves competitive accuracy across four benchmark datasets compared to MoE baselines. The core distinctions are summarized as follows:

    \begin{figure*}[t]
        \centering
        \includegraphics[width=0.8\textwidth]{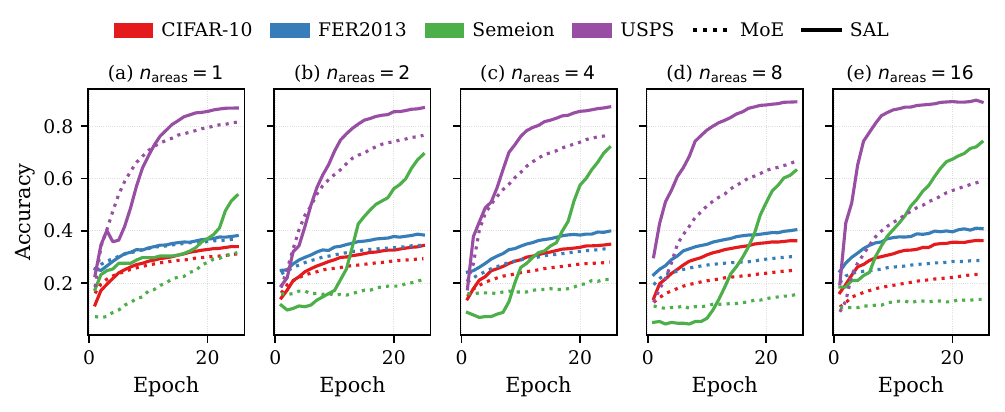}
        \caption{Accuracy comparison between SAL and MoE across various networks and 4 datasets. The MoE architecture follows the same structure with Table~\ref{tab:main_final_test_accuracy_sal_vs_baseline}, where the $n_{\text{areas}}$ parameter from SAL serves as the number of experts.}
        \label{fig:main_sal_with_moe_on_4_datasets}
    \end{figure*}

    \begin{itemize}
        \item \textbf{Decoupled Credit Assignment:} Unlike MoE's reliance on global backpropagation, SAL utilizes asymmetric alignment with independent feedback weights. This bypasses the weight transport problem and allows for more biologically plausible, local weight updates.
        \item \textbf{System-Wide Sparsity:} While MoE sparsity is often confined to specific layers (like FFNs) and requires complex load balancing, SAL enables end-to-end sparsity through fixed feedback pathways and explicit semantic partitioning across the entire architecture.
    \end{itemize}

    \subsection{Convergence Stability and Optimization}

    SAL achieves stable convergence by integrating fixed asymmetric feedback with local alignment signals. Unlike the unified gradient flow in BP, SAL employs a dual-signal mechanism: local terms provide direct objectives for intermediate layers, while global errors maintain overall task coherence.

    The area-partitioning mechanism, though non-differentiable due to the $\arg\max$ operator, is optimized through two complementary strategies:
    \begin{itemize}
        \item \textbf{Auxiliary Supervision:} Selector parameters are updated via a differentiable objective independent of the backbone gradients, ensuring stable routing.
        \item \textbf{Conditional Differentiability:} Once a path is selected, each active subnetwork remains differentiable, allowing for standard optimization post-routing.
    \end{itemize}
    By partitioning the latent space into discrete operational areas, SAL mitigates parameter contention and promotes a more orthogonal gradient flow, potentially facilitating smoother optimization on complex loss surfaces.

    \subsection{Limitations and Future Directions}

    Despite its advantages, several challenges remain for future investigation:
    \begin{itemize}
        \item \textbf{Memory Overhead:} The area-conditional parameterization leads to a linear increase in memory requirements relative to the number of areas ($n_{\text{areas}}$).
        \item \textbf{Routing Sensitivity:} Hard routing assignments are sensitive to initialization. Suboptimal early-stage partitioning may impede the convergence rate or lead to local optima.
        \item \textbf{Architectural Scalability:} While validated on feedforward and residual structures, extending SAL to more complex architectures like Transformers requires a careful balance between routing flexibility and computational efficiency.
    \end{itemize}

    \section{Conclusion}

    In this paper, we introduced the SAL method, a biologically motivated training approach designed to mitigate gradient interference and address the requirement for weight symmetry in neural networks. The method achieves this by selective parameter activation with adaptive area partitioning.

    Experimental results demonstrate that the SAL method maintains consistent convergence and stability across a range of model depths and parameter scales. These findings suggest that SAL exhibits improved robustness against overfitting compared to standard Backpropagation. Furthermore, SAL features an inherent sparsification mechanism, which potentially contributes to enhanced model generalization.

    Overall, SAL represents a continuing effort to incorporate biological principles into scalable artificial intelligence systems.

    \bibliography{main}
    \bibliographystyle{IEEEtran}

    \newpage
    \appendix
    \onecolumn

    \section{Complete Training Details for the 10 Datasets}
    \label{appendix:Complete Training Details for the 10 Datasets}
    While Table~\ref{tab:main_final_test_accuracy_sal_vs_baseline} reports validation accuracy, the training and validation losses are included in Table~\ref{tab:appendix_metics_table} to provide a more comprehensive overview.

    Across ten diverse datasets, the results demonstrate a consistent trend, supporting the effectiveness of the SAL method. Although these validations are currently focused on classification tasks, they provide a foundation for expansion into broader scenarios. Specifically, further verification regarding model depth and width—as discussed in Sections~\ref{lab:impact_of_network_width} and \ref{lab:impact_of_network_depth}—provides evidence for the reliability of the method in achieving convergence.

    \begin{table*}[t]
        \caption{Loss comparison between SAL and baseline methods across 10 benchmarks}
        \label{tab:appendix_metics_table}
        \centering

        \begin{subtable}{\textwidth}
            \centering
            \caption{Final training loss. A comparison of training loss trajectories reveals a clear distinction between the two approaches; the SAL method demonstrates superior convergence and better adaptation to the training set compared to the baseline.}
            \label{tab:appendix_metrics_table_part_1}
            \scshape
            \input{figures/appendix_final_train_loss_sal_vs_baseline}
        \end{subtable}
        \vspace{1.5em}

        \begin{subtable}{\textwidth}
            \centering
            \caption{Final validation loss. With the exception of the PCam dataset, the SAL method consistently yielded a lower loss than the baseline. Furthermore, a continuing downward trend in loss values was observed as $n\_areas$ increased.}
            \label{tab:appendix_metrics_table_part_2}
            \scshape
            \input{figures/appendix_final_test_loss_sal_vs_baseline}
        \end{subtable}
    \end{table*}

    \section{Supplementary Validation: Performance Comparison with MoE}
    \label{appendix:Performance Comparison with MoE}

    While the four datasets presented in Figure~\ref{fig:main_sal_with_moe_on_4_datasets} demonstrate that the effectiveness of SAL is not merely a coincidental result of partitioning data into specific areas, we extended our comparative experiments to the Fashion-MNIST, MNIST, PCam, and STL-10 datasets. These additional evaluations utilized the same network architectures to further substantiate the validity of our findings.

\begin{figure*}[t]
        \centering
        \includegraphics[width=0.8\textwidth]{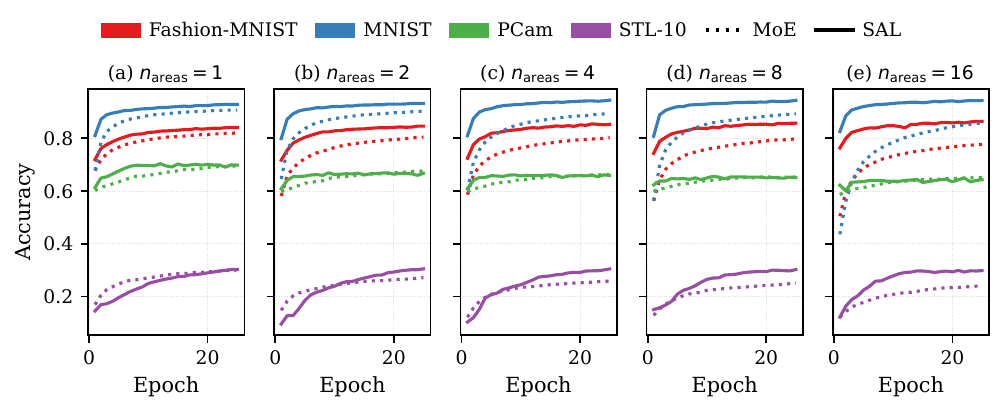}
        \caption{Acucracy comparison between SAL and MoE across various networks and 4 datasets.}
        \label{fig:appendix_sal_with_moe_on_4_datasets}
    \end{figure*}

    As illustrated in Figure~\ref{fig:appendix_sal_with_moe_on_4_datasets}, the overall performance ranking remains largely consistent as $n_{\text{areas}}$ increases from 1 to 16. Although the performance is relatively comparable on the PCam dataset, SAL consistently achieves higher final accuracy than the MoE approach across the other datasets. These results suggest that SAL possesses superior generalization capabilities and learning efficiency in feature extraction and multi-area task processing, further supporting our discussion in Section~\ref{sec:sal_vs_moe}.

    Notably, during the early stages of training, SAL exhibits a significantly faster increase in accuracy compared to MoE. This suggests that the selective parameter activation with adaptive area partitioning proposed in SAL accelerates the localization and learning of key features, thereby reducing the training time required to reach optimal performance.

\end{document}

%% file: figures/main_final_test_accuracy_sal_vs_baseline.tex
\begin{tabular}{lcccccc}
\toprule
Dataset & Baseline & SAL-1 & SAL-2 & SAL-4 & SAL-8 & SAL-16 \\
\midrule
CIFAR-10 & 30.81 $\pm$ 0.24 & 33.90 $\pm$ 0.26 & 34.46 $\pm$ 0.27 & 35.30 $\pm$ 0.39 & 36.20 $\pm$ 0.19 & \textbf{36.60 $\pm$ 0.48} \\
Digits & 38.53 $\pm$ 5.78 & 36.36 $\pm$ 3.04 & 35.11 $\pm$ 8.05 & 53.59 $\pm$ 8.37 & 65.65 $\pm$ 4.92 & \textbf{71.63 $\pm$ 5.75} \\
FER2013 & 36.51 $\pm$ 0.41 & 38.05 $\pm$ 0.25 & 38.67 $\pm$ 0.23 & 40.05 $\pm$ 0.24 & \textbf{40.73 $\pm$ 0.42} & 40.46 $\pm$ 0.20 \\
Fashion-MNIST & 81.76 $\pm$ 0.16 & 84.16 $\pm$ 0.11 & 84.92 $\pm$ 0.24 & 85.40 $\pm$ 0.21 & 85.60 $\pm$ 0.19 & \textbf{86.20 $\pm$ 0.37} \\
MNIST & 90.64 $\pm$ 0.18 & 93.02 $\pm$ 0.11 & 93.48 $\pm$ 0.23 & 94.17 $\pm$ 0.36 & 94.45 $\pm$ 0.17 & \textbf{94.71 $\pm$ 0.32} \\
PCam & 69.18 $\pm$ 0.27 & \textbf{69.52 $\pm$ 0.36} & 66.92 $\pm$ 0.34 & 65.44 $\pm$ 0.47 & 65.01 $\pm$ 0.61 & 64.83 $\pm$ 0.27 \\
STL-10 & 29.06 $\pm$ 0.12 & 29.82 $\pm$ 0.29 & 30.07 $\pm$ 0.49 & \textbf{30.42 $\pm$ 0.11} & 30.19 $\pm$ 0.30 & 29.01 $\pm$ 0.52 \\
SVHN & 47.08 $\pm$ 1.05 & 66.38 $\pm$ 0.32 & 66.73 $\pm$ 0.39 & 67.04 $\pm$ 0.46 & 66.85 $\pm$ 1.39 & \textbf{68.69 $\pm$ 1.26} \\
Semeion & 35.16 $\pm$ 2.72 & 51.00 $\pm$ 6.81 & 62.56 $\pm$ 6.36 & 65.20 $\pm$ 4.43 & 67.20 $\pm$ 3.30 & \textbf{72.03 $\pm$ 3.97} \\
USPS & 81.74 $\pm$ 1.12 & 87.07 $\pm$ 0.27 & 86.85 $\pm$ 0.52 & 87.90 $\pm$ 0.56 & 88.70 $\pm$ 0.51 & \textbf{89.36 $\pm$ 0.73} \\
\bottomrule
\end{tabular}

%% file: figures/appendix_final_train_loss_sal_vs_baseline.tex
\begin{tabular}{lcccccc}
\toprule
Dataset & Baseline & SAL-1 & SAL-2 & SAL-4 & SAL-8 & SAL-16 \\
\midrule
CIFAR-10 & 1.94 $\pm$ 0.00 & 1.90 $\pm$ 0.00 & 1.87 $\pm$ 0.00 & 1.81 $\pm$ 0.00 & 1.70 $\pm$ 0.00 & \textbf{1.53 $\pm$ 0.01} \\
Digits & 1.81 $\pm$ 0.14 & 2.22 $\pm$ 0.01 & 2.19 $\pm$ 0.02 & 2.06 $\pm$ 0.08 & 1.84 $\pm$ 0.02 & \textbf{1.57 $\pm$ 0.12} \\
FER2013 & 1.62 $\pm$ 0.00 & 1.59 $\pm$ 0.00 & 1.54 $\pm$ 0.01 & 1.42 $\pm$ 0.01 & 1.22 $\pm$ 0.01 & \textbf{0.97 $\pm$ 0.01} \\
Fashion-MNIST & 0.49 $\pm$ 0.00 & 0.41 $\pm$ 0.00 & 0.39 $\pm$ 0.00 & 0.35 $\pm$ 0.00 & 0.32 $\pm$ 0.00 & \textbf{0.28 $\pm$ 0.00} \\
MNIST & 0.35 $\pm$ 0.00 & 0.26 $\pm$ 0.00 & 0.23 $\pm$ 0.01 & 0.19 $\pm$ 0.01 & 0.16 $\pm$ 0.00 & \textbf{0.14 $\pm$ 0.00} \\
PCam & 0.45 $\pm$ 0.01 & 0.16 $\pm$ 0.01 & 0.11 $\pm$ 0.01 & 0.09 $\pm$ 0.01 & \textbf{0.08 $\pm$ 0.01} & 0.10 $\pm$ 0.01 \\
STL-10 & 1.78 $\pm$ 0.01 & 1.81 $\pm$ 0.01 & 1.67 $\pm$ 0.01 & 1.38 $\pm$ 0.02 & 0.95 $\pm$ 0.02 & \textbf{0.56 $\pm$ 0.02} \\
SVHN & 1.76 $\pm$ 0.02 & 1.17 $\pm$ 0.00 & 1.15 $\pm$ 0.02 & 1.12 $\pm$ 0.03 & 1.10 $\pm$ 0.05 & \textbf{0.95 $\pm$ 0.05} \\
Semeion & 1.86 $\pm$ 0.09 & 1.92 $\pm$ 0.03 & 1.77 $\pm$ 0.08 & 1.59 $\pm$ 0.06 & 1.40 $\pm$ 0.10 & \textbf{1.01 $\pm$ 0.06} \\
USPS & 0.71 $\pm$ 0.04 & 0.48 $\pm$ 0.01 & 0.43 $\pm$ 0.02 & 0.38 $\pm$ 0.01 & 0.31 $\pm$ 0.01 & \textbf{0.25 $\pm$ 0.01} \\
\bottomrule
\end{tabular}

%% file: figures/appendix_final_test_loss_sal_vs_baseline.tex
\begin{tabular}{lcccccc}
\toprule
Dataset & Baseline & SAL-1 & SAL-2 & SAL-4 & SAL-8 & SAL-16 \\
\midrule
CIFAR-10 & 1.96 $\pm$ 0.00 & 1.91 $\pm$ 0.00 & 1.90 $\pm$ 0.00 & 1.88 $\pm$ 0.00 & 1.84 $\pm$ 0.01 & \textbf{1.82 $\pm$ 0.01} \\
Digits & 1.81 $\pm$ 0.12 & 2.22 $\pm$ 0.01 & 2.20 $\pm$ 0.02 & 2.07 $\pm$ 0.08 & 1.85 $\pm$ 0.04 & \textbf{1.63 $\pm$ 0.13} \\
FER2013 & 1.65 $\pm$ 0.00 & 1.62 $\pm$ 0.00 & 1.60 $\pm$ 0.00 & 1.57 $\pm$ 0.01 & \textbf{1.56 $\pm$ 0.01} & 1.58 $\pm$ 0.01 \\
Fashion-MNIST & 0.52 $\pm$ 0.00 & 0.45 $\pm$ 0.00 & 0.43 $\pm$ 0.00 & 0.41 $\pm$ 0.00 & 0.40 $\pm$ 0.00 & \textbf{0.38 $\pm$ 0.01} \\
MNIST & 0.33 $\pm$ 0.00 & 0.25 $\pm$ 0.00 & 0.23 $\pm$ 0.01 & 0.21 $\pm$ 0.01 & 0.19 $\pm$ 0.00 & \textbf{0.18 $\pm$ 0.01} \\
PCam & \textbf{0.58 $\pm$ 0.00} & 0.65 $\pm$ 0.01 & 0.78 $\pm$ 0.02 & 0.92 $\pm$ 0.04 & 1.04 $\pm$ 0.08 & 1.26 $\pm$ 0.08 \\
STL-10 & 1.95 $\pm$ 0.01 & 1.91 $\pm$ 0.01 & 1.89 $\pm$ 0.01 & \textbf{1.89 $\pm$ 0.01} & 1.96 $\pm$ 0.01 & 2.11 $\pm$ 0.01 \\
SVHN & 1.73 $\pm$ 0.02 & 1.22 $\pm$ 0.00 & 1.21 $\pm$ 0.01 & 1.19 $\pm$ 0.02 & 1.19 $\pm$ 0.05 & \textbf{1.10 $\pm$ 0.05} \\
Semeion & 1.89 $\pm$ 0.07 & 1.92 $\pm$ 0.03 & 1.78 $\pm$ 0.08 & 1.62 $\pm$ 0.05 & 1.47 $\pm$ 0.11 & \textbf{1.18 $\pm$ 0.06} \\
USPS & 0.79 $\pm$ 0.04 & 0.57 $\pm$ 0.01 & 0.53 $\pm$ 0.01 & 0.49 $\pm$ 0.02 & 0.44 $\pm$ 0.01 & \textbf{0.40 $\pm$ 0.01} \\
\bottomrule
\end{tabular}